%% file: arxiv.tex
\def\mytitle{Leverage Score Sampling for\\ Faster Accelerated Regression and ERM}

\def\myauthors{Naman Agarwal\footnote{namana@cs.princeton.edu,
Computer Science, Princeton University},\hspace{0.7cm} Sham Kakade \footnote{sham@cs.washington.edu,
University of Washington, Seattle}, \hspace{0.7cm}Rahul Kidambi \footnote{rkidambi@uw.edu, University of Washington, Seattle}, \hspace{0.7cm}Yin Tat Lee \footnote{yintat@uw.edu, University of Washington, Seattle}, \\ \vspace{0.4cm} Praneeth Netrapalli, \footnote{praneeth@microsoft.com, Microsoft Research, India}, Aaron Sidford, \footnote{sidford@stanford.edu, Stanford University}}

\def\myappendix{Appendix}
\documentclass[11pt]{article}
\usepackage[fullpage,nousetoc]{paper}
\usepackage{algorithm,algorithmicx,algpseudocode}
\usepackage[lined,boxed,ruled,norelsize,algo2e]{algorithm2e}

\theoremstyle{plain}

\interfootnotelinepenalty=10000

\include{general_notation}


\numberwithin{equation}{section}
\numberwithin{figure}{section}
\theoremstyle{plain}
\newtheorem{thm}{\protect\theoremname}
\theoremstyle{definition}
\newtheorem{defn}[thm]{\protect\definitionname}
\theoremstyle{corollary}

\theoremstyle{remark}

\theoremstyle{remark}
\newtheorem*{rem*}{\protect\remarkname}
\theoremstyle{plain}
\newtheorem{lem}[thm]{\protect\lemmaname}
\theoremstyle{assumption}

\providecommand{\definitionname}{Definition}
\providecommand{\lemmaname}{Lemma}
\providecommand{\remarkname}{Remark}
\providecommand{\theoremname}{Theorem}
\providecommand{\corollaryname}{Corollary}
\providecommand{\assumptionname}{Assumption}

\newcommand{\condtotal}{\kappa_{\mathrm{sum}}}

\global\long\def\alg{\mathcal{A}}

\def\ermname{ERM }

\BEGINDOC
\begin{titlepage}
\makeheader
\begin{abstract}
Given a matrix $\ma\in\R^{n\times d}$ and a vector $\rhs \in\R^{d}$,
we show how to compute an $\epsilon$-approximate solution to the regression problem 
$
\min_{x\in\R^{d}}\frac{1}{2} \norm{\ma x - \rhs}_{2}^{2}
$
in time $
\otilde ((n+\sqrt{d\cdot\kappa_{\text{sum}}})\cdot s\cdot\log\epsilon^{-1})
$
where $\kappa_{\text{sum}}=\tr\left(\ma^{\top}\ma\right)/\lambda_{\min}(\ma^{T}\ma)$
and $s$ is the maximum number of non-zero entries in a row of $\ma$. Our algorithm improves upon the previous best running time of 
$
\otilde ((n+\sqrt{n \cdot\kappa_{\text{sum}}})\cdot s\cdot\log\epsilon^{-1})$.

We achieve our result through a careful combination of leverage score sampling techniques, proximal point methods, and accelerated coordinate descent. Our method not only matches the performance of previous methods, but further improves whenever leverage scores of rows are small (up to polylogarithmic factors). We also provide a non-linear generalization of these results that improves the running time for solving a broader class of ERM problems. 
\end{abstract} 
\end{titlepage}
\pagebreak

\section{Introduction}

Given $\ma\in\R^{n\times d}$ and $\rhs \in \R^{n}$,  the regression
problem 
$
\min_{x\in\R^{d}}\frac{1}{2}\norm{\ma x - \rhs}_{2}^{2}
$
is one
of the most fundamental problems in optimization and a prominent tool in machine learning. It is one of the simplest empirical risk minimization (ERM) problems and a prominent proving ground for developing efficient learning algorithms. 

Regression is long known to be solve-able directly by matrix multiplication
in $O(nd^{\omega-1})$ time where $\omega < 2.373$ \cite{Williams12} is the matrix multiplication constant and recent work
has improved the running time to 
$
\otilde((\nnz(\ma)+d^{\omega})\log(\epsilon^{-1}))
$,\footnote{Through this paper, we use $\otilde$ to hide factors polylogarithmic
	in $n$,$d$,$\kappa\defeq\lambda_{\max}(\ma^{\top}\ma)/\lambda_{\min}(\ma^{\top}\ma)$ and $M$ (c.f. Definition~\ref{def:regularity} and Definition~\ref{defn:ERMdef}).}
i.e. linear time plus the time needed to solve
a nearly square linear system. \cite{ClarksonW13,iterativeRowSampling,osnap,uniformSampling,cohenSubspace} However, for sufficiently large $\ma$ even a super-quadratic running time of $\Omega(d^{\omega})$ \ can be prohibitively
expensive. Consequently, over the past decade improving this running time under mild regularity assumptions on $\ma$ has been an incredibly active area of research. 

In this paper we improve the best known running time for solving regression under standard regularity assumptions. Formally the problem we consider is as follows.

\begin{defn}[The Regression Problem]
	\label{def:regression} Given $\ma\in\R^{n\times d}$ with rows
	$a_{1},...,a_{n}$ and $\rhs \in\R^{n}$, we consider the \emph{regression
		problem} $\min_{x\in\R^{d}}f_{\ma,\rhs}(x)$ where 
	\[
	f_{\ma,\rhs}(x)\defeq\frac{1}{2}\norm{\ma x-\rhs}_{2}^{2}=\sum_{i\in[n]}\frac{1}{2}\left(a_{i}^{\top}x-\rhs_{i}\right)^{2}\,.
	\]

\end{defn}

The central problem of this paper is to get faster regression algorithms defined as follows. 

\begin{defn}[Regression Algorithm]
\label{def:regress_alg}
We call an algorithm a \emph{$\mathcal{T}(\ma)$-time
	regression algorithm} if for any $\rhs\in\R^{n}$, $x_{0}\in\R^{d}$, and $\epsilon \in (0, \frac{1}{2})$ w.h.p in $n$ in time $O(\mathcal{T}(\ma) \log \epsilon^{-1})$
the algorithm outputs a vector $y$ such that
\begin{equation}
f_{\ma,\rhs}(y)-\min_{x}f_{\ma,\rhs}(x)\leq\epsilon\cdot\left(f_{\ma,\rhs}(x_{0})-\min_{x}f_{\ma,\rhs}(x)\right)\,.\label{eq:fval_guarantee}
\end{equation}
\end{defn}

Note that if $x_{*}$ is a minimizer of $f_{\ma,\rhs}(x)$ then the guarantee
(\ref{eq:fval_guarantee}) is equivalent to the following 
\begin{equation}
\norm{ y- x_{*}}_{\ma^\top \ma}^{2}\leq\epsilon\norm{ x_{0} - x_{*}}_{\ma^\top \ma}^{2}\label{eq:dist_guarantee}
\end{equation}
where $\|x\|_{\mm}^2$ for a PSD matrix $M$ is defined as $x^{\top}\mm x$.

The goal of this paper is to provide regression algorithms with improved running times depending on $n$, $d$, and the following regularity parameters. 

\begin{defn} (Regularity Parameters) 
\label{def:regularity}
We let $\lambda_{\min}(\ma^{\top}\ma)$ and $\lambda_{\max}(\ma^{\top}\ma)$ denote the smallest and largest eigenvalues of $\ma^{\top}\ma$. We let $\kappa(\ma^{\top}\ma)=\lambda_{\max}(\ma^{\top}\ma) / \lambda_{\min}(\ma^{T}\ma)$ denote the \emph{condition number} of $\ma^\top \ma$ and let  $\condtotal(\ma^{\top}\ma)= \tr\left(\ma^{\top}\ma\right) / \lambda_{\min}(\ma^{T}\ma)$ denote the \emph{total condition number} of $\ma^\top \ma$. We let $s(\ma)$ denote the maximum number of non-zero entries in a row of $\ma$. Occasionally, we drop the terms in parenthesis when they are clear from context.
%
%
%
\end{defn}

\subsection{Previous Results}

Standard classic iterative methods such as gradient descent and accelerated
gradient descent \cite{Nesterov1983} solve the regression problem with running times of 
$
O(n\cdot s(\ma) \cdot \kappa(\ma^\top \ma))
$
and 
$
O (n\cdot s(\ma) \cdot \sqrt{\kappa(\ma^\top \ma)} )
$
respectively. While these running times are super-linear whenever $\kappa(\ma^\top \ma)$ is super constant there has been a flurry of recent papers showing that using sampling techniques faster running times can be achieved. These often yield nearly linear running times when $n$ is sufficiently larger than $d$. \cite{sdca,Johnson013,accelSDCA,katyusha}. 

Using recent advances in accelerated coordinate descent \cite{accelerationNonUniform,Ns16} coupled with proximal point methods \cite{proximalPoint,catalyst} 
the previous fastest iterative algorithm is as follows:

\begin{thm}[Previous Best Regression Running Times]
	\label{thm:standard_solvers} Given $\ma\in\R^{n\times d}$, there is a $\mathcal{T}(\ma)$-time regression
	algorithm with
	\[
\runtime(\ma) = 
	\otilde\left(\left(n + \frac{\sum_{i\in[n]}\norm{a_{i}}_{2}}{\sqrt{\lambda_{\min} (\ma^\top \ma )}}\right)\cdot s(\ma)\right) = \otilde((n+\sqrt{n\cdot\condtotal(\ma^\top \ma)})\cdot s(\ma )).
	\]
\end{thm}

The inequality in this theorem follows directly from Cauchy Schwartz, as 
\[
\sum_{i\in[n]}\norm{a_{i}}_{2}\leq\sqrt{n\cdot\tr(\ma^{\top}\ma)}=\sqrt{n\cdot\condtotal(\ma^\top \ma)\cdot\lambda_{\min}(\ma^\top \ma)}
~.
\]
For the rest of the proof, see Section~\ref{sec:Deferred}, where we provide a proof of Theorem~\ref{thm:standard_solvers2}, a generalization of Theorem~\ref{thm:standard_solvers} which is more convenient for our analysis.

%
%
%
%
%

\subsection{Our Results}

The work in this paper is motivated by the natural question, 
\emph{can this running time of Theorem~\ref{thm:standard_solvers} be further improved}?
Despite the running time lower bound of $\sqrt{n\cdot\condtotal(\ma^{\top}\ma)}$ shown in \cite{woodworth2016tight},\footnote{Their lower bound involves a function with $d\gg n$. However, $d\ll n$ is more common as we explain.} in this paper we give an affirmative answer improving the 
$\sqrt{n\cdot\condtotal(\ma^{\top}\ma)}$ term in Theorem~\ref{thm:standard_solvers} to $\sqrt{d\cdot\condtotal(\ma^{\top}\ma)}$. The main result of this paper is the following:
\ifdefined\isaccepted
\footnote{A preliminary version of the work in this paper was referenced in \cite{agarwal2016second}, which gave a subset of the results in this paper.}
\fi

\begin{thm}[Improved Regression Running Time]
\label{thm:our_solvers_fine} 
Given $\ma\in\R^{n\times d}$,
	Algorithm~\ref{alg:solveUsingLS} is a $\runtime(\ma)$-time
	regression algorithm that succeeds with high probability(w.h.p) in $n$ where 
\[
\runtime(\ma) =
\otilde \left(\nnz(\ma)+\left(d+\frac{\sum_{i\in[n]}\norm{a_{i}}_2 \cdot \sigma_i(\ma)}{\sqrt{\lambda_{\min}(\ma^{T}\ma)}}\right) \cdot s(\ma) \right)
\]
and $\sigma_i(\ma) = \norm{a_{i}}^2_{(\ma^{\top}\ma)^{-1}}$ for all $i \in [n]$.
\end{thm}
Up to polylogarithmic factors Theorem~\ref{thm:our_solvers_fine}
is an improvement over Theorem~\ref{thm:standard_solvers} as $\sigma_i(\ma) \in [0, 1] $.
This improvement can be substantial as $\sigma_i(\ma)$ can be as small as $O(d/n)$, e.g. if $\ma$ is an entry-wise random Gaussian matrix. Compared to Theorem~\ref{thm:standard_solvers}
whose second term in running time grows as $n$, our second term is
independent of $n$ due to the following: 
\begin{equation}
\label{eqn:levscorered}
\sum_{i\in[n]}\norm{a_{i}}_2 \sigma_i(\ma)
 \leq\sqrt{\sum_{i\in[n]}\norm{a_{i}}_{2}^{2}\sum_{i\in[n]}\norm{a_{i}}_{(\ma^{\top}\ma)^{-1}}^{2}} = \sqrt{\tr\left(\ma^{\top}\ma\right)\tr(\ma(\ma^{\top}\ma)^{-1}\ma^{T})}
\leq\sqrt{d \kappa_{\text{sum}}}.
\end{equation}
Therefore in Theorem~\ref{thm:our_solvers_fine} we have 
$
\runtime(\ma) = \otilde((n + \sqrt{d \cdot \condtotal(\ma^\top \ma)})\cdot s(\ma) )$.

This improvement from $n$ to $d$ can be significant as 
$n$ (the number of samples) is in some cases orders of magnitude larger than $d$ (the number of features).
For example, in the LIBSVM dataset\footnote{https://www.csie.ntu.edu.tw/\textasciitilde{}cjlin/libsvmtools/datasets/},
in 87 out of 106 many non-text problems, we have $n\geq d$, 50
of them have $n\geq d^{2}$ and in the UCI dataset,\footnote{http://archive.ics.uci.edu/ml/datasets.html}
in 279 out of 301 many non-text problems, we have $n\geq d$, 195
out of them have $n\geq d^{2}$.

Furthermore, in Section~\ref{sec:erm} we show how to extend our results to ERM problems more general then regression. In particular we consider the following ERM problem

\begin{defn}[ERM]
\label{defn:ERMdef}
Given $\ma\in\R^{n\times d}$ with rows
	$a_{1},...,a_{n}$ and functions $\{\psi_1 \ldots \psi_n\} \in \R \rightarrow \R$ such that each $\psi_{i}:\R\rightarrow\R$ is twice differentiable and satisfies
	\begin{equation}
	 \label{eqn:generalERMEquation}	
	 \forall\;x \in \R^d\;\;\; \frac{1}{M} \leq \psi''(x) \leq M
	 \end{equation}
	 we wish to minimize $F(x) : \R^{d} \rightarrow \R$ over $x \in \R^d$ where 
\[F(x) \defeq \sum_{i\in[n]}f_i(x) = \sum_{i\in[n]}\psi_{i}(a_i^Tx)\]
\end{defn}

While the form of ERM considered by us is not the most general form considered in literature, we consider this problem, to provide a proof of concept that our techniques for regression are more broadly applicable to the case of ERM and can if fact provide a non-trivial improvement over existing methods.  We leave it to future work to study the limitations of our techniques and possible improvements for more general ERM problems. The following is our main theorem regarding the ERM problem. 

\begin{thm}
\label{thm:mainermtheorem}
	Given an ERM problem(Definition \ref{defn:ERMdef}) and an initial point $x_0$, there exists an algorithm that produces a point $x'$ such that
	$ 
	F(x') - F_{x_*} \leq \epsilon \left(F(x_0) - F_{x_*} \right)$ 
	which succeeds w.h.p in n in total time
	\[ \tilde{O}\left(\left( \nnz(\ma) + \left( d M^5 + \sum_{i=1}^{n} \frac{\|a_{i}\|_2 \sqrt{\sigma_i}M^3}{ \sqrt{\lambda_{\min}(\ma^\top \ma)}} \right)  s(\ma)\right)\log\left(\frac{1}{\epsilon}\right)\right)\]
	where $\sigma_i \defeq a_i^T [\ma^{\top} \ma]^{-1} a_i$ are the leverage scores with respect to $a_i$.
\end{thm}

Note that Theorem \ref{thm:mainermtheorem} interpolates our regression results, i.e. it recovers our results for regression in the special case of $M = 1$. To better understand the bound in Theorem~\ref{thm:mainermtheorem}, note that following the derivation in Equation~\eqref{eqn:levscorered} we have that the running time in Theorem~\ref{thm:mainermtheorem} is bounded by

\[ \tilde{O}\left(\left( \nnz(\ma) + \left( dM^5 + \sum_{i=1}^{n} M^3\sqrt{d \condtotal(\ma^{\top}\ma)}\right)  s(\ma)\right)\log\left(\frac{1}{\epsilon}\right)\right) ~. \]
The best known bound for the ERM problem as defined in Definition \ref{defn:ERMdef} given by \cite{proximalPoint,catalyst,katyusha} is 
\[ \tilde{O}\left(\left( \nnz(\ma) + \sum_{i=1}^{n} M\sqrt{n\condtotal(\ma^{\top}\ma))}\right)  s(\ma)\log\left(\frac{1}{\epsilon}\right)\right) \]
In this case Theorem \ref{thm:mainermtheorem} should be seen as implying that under Assumption \eqref{eqn:generalERMEquation} the effective dependence on the number of examples on the running time for ERM can be reduced to at most $dM^5$. 

Again, we remark that the running time bound of Theorem~\ref{thm:mainermtheorem} should be viewed as a proof of concept that our regression machinery can be used to improve the running time of ERM. We leave it as future work to both improve Theorem~\ref{thm:mainermtheorem}'s dependence on $M$ and have it extend to a broader set of problems. For example, we believe the the running time can be immediately improved to 
 \[ \tilde{O}\left(\left( \nnz(\ma) + \left( dM^4 + \sum_{i=1}^{n} \frac{\|a_{i}\|_2\sqrt{\sigma_i}M^3}{ \sqrt{\lambda_{\min}(\ma^\top \ma)}} \right)  s(\ma)\right)\log\left(\frac{1}{\epsilon}\right)\right)\]
 simply by using a proximal version of Theorem~\ref{thm:standard_solvers_gen_ERM}, which is alluded to in the work of \cite{accelerationNonUniform}. Note that this improvement leads to the effective number of examples being bounded by $dM^4$.

\subsection{Our Approach}

Our algorithm follows from careful combination and analysis of a recent suite of advances in numerical linear algebra. 
First, we use the previous fastest regression algorithm, Theorem~\ref{thm:standard_solvers}, which is the combination of recent advances in accelerated coordinate descent \cite{accelerationNonUniform,Ns16} and proximal point methods \cite{proximalPoint,catalyst} (See Section~\ref{sec:Deferred}.) 
Then, we show that if we have estimates of the leverage scores of the rows of $\ma$, a natural recently popularized measure of importance, \cite{spielmanSrivastava,iterativeRowSampling,uniformSampling} we can use concentration results on leverage score sampling and preconditioning to obtain a faster regression algorithm. (See Section~\ref{sec:alg_ls}.)

Now, it is a powerful and somewhat well known fact that given an algorithm for regression one can compute leverage score estimates in nearly linear time plus the time needed to solve $\otilde(1)$ regression problems \cite{spielmanSrivastava}. Consequently, to achieve the improved running time when we do not have leverage scores we are left with a chicken and egg problem. Fortunately, recent work has shown that such a problem can be solved in a several ways \cite{iterativeRowSampling,uniformSampling}. We show that the technique in \cite{iterativeRowSampling} carefully applied and analyzed can be used to obtain our improved running time for both estimating leverage scores and solving regression problems with little overhead. (See Section~\ref{sec:alg_no_ls}.)

To generalize our results for a broader class of ERM problems we follow a similar procedure. Most parts generalize naturally, however perhaps the most complex ingredient is how to generalize preconditioning to the case when we are sampling non-quadratic functions. For this, we prove concentration results on sampling from ERM inspired from \cite{frostig2015competing} to show that it suffices to solve ERM on a sub-sampling of the components that may be of intrinsic interest. (See Section~\ref{sec:erm})

In summary our algorithms are essentially a careful blend of accelerated coordinate descent and concentration results coupled with the iterative procedure in \cite{iterativeRowSampling} and the Johnson Lindenstrauss machinery of \cite{spielmanSrivastava} to compute leverage scores. Ultimately the algorithms we provide are fairly straightforward, but it provides a substantial running time improvement that we think is of intrinsic interest.  We hope this work may serve as a springboard for even faster iterative methods for regression and minimizing finite sums more broadly.

Finally, we remark that there is another way to achieve the $\sqrt{d \cdot \condtotal(\ma)}$ improvement over $\sqrt{n \cdot \condtotal(\ma)}$. One could simply use subspace embeddings \cite{ClarksonW13,osnap,cohenSubspace} and preconditioning to reduce the regression problem to a regression problem on a $\otilde(d) \times d$ matrix and then apply Theorem~\ref{thm:standard_solvers} to solve the $\otilde(d) \times d$ regression problem. While this works, it has three shortcomings relevant to our approach. First, this procedure would possibly lose sparsity, the $\otilde(d) \times d$ matrix may be dense and thus the final running time would have an additional $\otilde(d^2)$ term our method does not. Second, it is unclear if this approach yields our more fine-grained  running time dependence on leverage scores that appears in Theorem~\ref{thm:our_solvers_fine} which we believe to be significant. Thirdly it is unclear how to extend the approach to the ERM setting. 

\subsection{Paper Organization}

After providing requisite notation in Section \ref{sec:notation} we prove Theorem \ref{thm:our_solvers_fine} in Sections \ref{sec:alg_ls} and \ref{sec:alg_no_ls}. We first provide the algorithm for regression given leverage score estimates in Section \ref{sec:alg_ls} and further provide the algorithm to compute the estimates based in Section \ref{sec:alg_no_ls}. Note that the algorithm for computing leverage scores makes use of the algorithm for regression given leverage scores as a sub-routine. In Section \ref{sec:Deferred} we provide the proofs of deferred lemmas and theorems from Sections \ref{sec:alg_ls}, \ref{sec:alg_no_ls}. In Section \ref{sec:erm} we provide the proof of Theorem \ref{thm:mainermtheorem}. Further we collect some simple re-derivations and reductions from known results required through the paper in the Appendix.

\section{Notation}
\label{sec:notation}
For symmetric matrix $\mm\in\R^{d\times d}$ and $x\in\R^{d}$ we
let $\norm x_{\mm}^{2}=x^{\top}\mm x$. For symmetric matrix $\mn\in\R^{d\times d}$
we use $\mm\preceq\mn$ to denote the condition that $x^{\top}\mm x\leq x^{\top}\mn x$
for all $x\in\R^{d}$ and we define $\prec$, $\succeq$, and $\succ$
analogously. 
We use $\nnz(\ma)$ to denote the number of non-zero entries in $\ma$ and for a vector
$b\in\R^{n}$ we let $\nnz(b)$ denote the number of nonzero entries
in $b$. 


\section{Regression Algorithm Given Leverage Score Estimates}
\label{sec:alg_ls}

The regression algorithm we provide in this paper involves two steps. First we find which rows of $\ma$ are important, where importance is measured in terms of \emph{leverage score}. Second, we use these leverage scores to sample the matrix and solve the regression problem on the sampled matrix using Theorem \ref{thm:standard_solvers2}. In this section we introduce leverage scores and provide and analyze the second step of our algorithm.
\begin{defn}[Leverage Score]
\label{defn:levscores}
	For $\ma\in\R^{n\times d}$ with rows $a_{1},...,a_{n}\in\R^{d}$ we denote the \emph{leverage score of row $i\in[n]$} by $\sigma_{i}(\ma)\defeq a_{i}^{\top}\left(\ma^{\top}\ma\right)^{+}a_{i}$.
\end{defn}
Leverage score have numerous applications and are well studied. It
is well known that $\sigma_{i}(\ma)\in(0,1]$ for all $i\in[n]$ and
$\sum_{i\in[n]}\sigma_{i}(\ma)=\mathrm{rank}(\ma)$. The critical
fact we used about leverage scores is that sampling rows of $\ma$
according to any overestimate of leverage scores yields a good approximation
to $\ma$ after appropriate re-scaling \cite{uniformSampling,spielmanSrivastava}:

\begin{lem}[Leverage Score Sampling (Lemma~4 of \cite{uniformSampling}) ]
	\label{lem:spectral_sparsification} Let $\ma\in\R^{n\times d}$,
	let $\delta\in(0,\frac{1}{2})$, and let $u\in\R^{n}$ be overestimates
	of the leverage scores of $\ma$; i.e. $u_{i}\geq\sigma_{i}(\ma)$
	for all $i\in[n]$. Define 
$
p_{i}\defeq\min\left\{ 1,k \delta^{-2}u_{i}\log n\right\}
$
	for a sufficiently large absolute constant $k>0$ and let $\mh\in\R^{n\times n}$
	be a random diagonal matrix where independently $\mh_{ii}=\frac{1}{p_{i}}$
	with probability $p_{i}$ and $\mh_{ii}=0$ otherwise. With high
	probability in $n$, $\nnz(\mh)=O(d\cdot\delta^{-2}\cdot\log n)$ and
$
	(1-\delta)\cdot\ma^{\top}\ma\preceq\ma^{\top}\mh\ma\preceq(1+\delta)\cdot\ma^{\top}\ma $.

\end{lem}
\begin{algorithm2e}[h]
\caption{$\mathtt{SolveUsingLS}_{\ma,u}(x_{0},\rhs,\epsilon)$}
\label{alg:solveUsingLS}
	
\SetAlgoLined
	
	Let $p_{i}=\min\left\{ 1,k'\cdot u_{i}\log n\right\} $ where $k'$
	is a sufficiently large absolute constant.

	\Repeat{ 
$\sum_{i\in[n]}\norm{\precondrow_i}_{2} \leq 2\cdot\sum_{i\in[n]}\sqrt{k' \cdot u_{i}\log n}\cdot\norm{a_{i}}_{2}$
	}{
		Let $\mh\in\R^{n\times n}$ be a diagonal matrix where independently
		for all $i\in[n]$ we let $\mh_{ii} = \frac{1}{p_i}$ with probability $p_i$ and $0$ otherwise.
		
	Let $\precond=\sqrt{\mh}\ma$.
	}
	Invoke Theorem~\ref{thm:standard_solvers2} on $\ma$ and $\precond$ to
	find $y$ such that
	\[
	f_{\ma,\rhs}(y)-\min_{x}f_{\ma,\rhs}(x)\leq\epsilon\cdot\left(f_{\ma,\rhs}(x_{0})-\min_{x}f_{\ma,\rhs}(x)\right).
	\]

	\textbf{Output:} $y$.
	
\end{algorithm2e}
\begin{thm}
\label{thm:solver_using_leverage_scores} 
If $u\in\R^{n}$ satisfies
$
\sigma_{i}(\ma)\leq u_{i}\leq4\cdot\sigma_{i}(\ma)+
[n \cdot \kappa(\ma^{\top}\ma)]^{-1}
$
for all $i\in[n]$ then $\mathtt{SolveUsingLS}_{\ma,u}$ is a $\runtime(\ma)$-time regression algorithm where
\[
\mathcal{T}(\ma) = 
\otilde\left(\nnz(\ma)+\left(d+\frac{\sum_{i\in[n]}\sqrt{\sigma_{i}(\ma)}\cdot\norm{a_{i}}_{2}}{\sqrt{\lambda_{\min}(\ma^{\top}\ma)}}\right)\cdot s(\ma)\right) ~.
\]
\end{thm}
\begin{proof}
	Let $k'=\delta^{-2}\cdot k$ for $\delta=\frac{1}{10}$. Applying
	Lemma~\ref{lem:spectral_sparsification} yields with high probability in $n$ that 
	\begin{equation}
	\left(\frac{5}{6}\right) \ma^{\top}\ma\preceq\ma^{\top}\mh\ma\preceq \left(\frac{6}{5}\right) \ma^{\top}\ma\label{eq:AA_apx}
	\end{equation}
	where
$
	\ma^{\top}\mh\ma=\sum_{i\in[n]\,:\,\mh_{ii}\neq0}\precondrow_{i}\precondrow_{i}^{\top} ~,
$
 $\precondrow_{i}\defeq\frac{1}{\sqrt{p_{i}}}a_{i}$
	and $p_{i}\defeq\min\left\{ 1,k' \cdot u_{i}\log n\right\} $. Note that 
	\[
	\E \left[\sum_{i\in[n]}\norm{\precondrow_{i}}_{2} \right] =\sum_{i\in[n]}\frac{p_{i}}{\sqrt{p_{i}}}\norm{a_{i}}_{2}\leq\sum_{i\in[n]}\sqrt{k' u_{i}\log n} \norm{a_{i}}_{2} ~.
	\]
	Consequently, by Markov's inequality
\[
	\sum_{i\in[n]}\norm{\precondrow_{i}}_{2}\leq2\cdot\sum_{i\in[n]}\sqrt{k'\cdot u_{i}\log n}\cdot\norm{a_{i}}_{2}
\]
	with probability at least $1/2$ and the loop in the algorithm terminates with high
	probability in $n$ in $O(\log n)$ iterations. Consequently, the
	loop takes only $O(\nnz(\ma)+n\log n)$-time and since we only sampled
	$O(\log n)$ many independent copies of $\ma^{\top}\mh\ma$, the guarantee
	(\ref{eq:AA_apx}) again holds with high probability in $n$. 
	
	Using the guarantee (\ref{eq:AA_apx}) and Theorem~\ref{thm:standard_solvers2}
	on $\ma$ and $\precond\defeq\sqrt{\mh}\ma$, we can produce a $y$ we
	need in time $O(\log(\epsilon^{-1}))$ times
\[ 
\otilde\left(\left(\text{nnz}(\ma)+\left(d\log n+\frac{1}{\sqrt{\lambda}}\sum_{i\in[n]}\norm{\precondrow_{i}}_{2}\right)\cdot s(\ma)\right)\right)
\]
where we used that $\precond$ has at most $O(d\log n)$ rows with
	high probability in $n$. Since we know
\[
	\sum_{i\in[n]}\norm{\precondrow_{i}}_{2}\leq2\sum_{i\in[n]}\sqrt{k' \cdot u_{i}\log n}\cdot\norm{a_{i}}_{2},
\]
	all that remains is to bound $\sum_{i\in[n]}\sqrt{u_{i}}\norm{a_{i}}_2$. 
 However, $\ma^{\top}\ma\preceq\lambda_{\max}(\ma^{\top}\ma)\mI$ and therefore
\[
\mI\preceq\lambda_{\max}(\ma^{\top}\ma)(\ma^{\top}\ma)^{-1}
\text{ and }
\norm{a_{i}}_{2}\leq\sqrt{\lambda_{\max}(\ma^{\top}\ma)}\cdot\sigma_{i}(\ma) ~.
\] 
Consequently, Cauchy Schwartz and  $\lambda_{\min}(\ma^{\top}\ma)\leq\tr(\ma^{\top}\ma)$ yields
\begin{equation*}
\frac{1}{\sqrt{n}}\sum_{i\in[n]}\norm{a_{i}}_2 \leq \sqrt{\sum_{i\in[n]}\norm{a_{i}}^{2}_2}  \leq\frac{1}{\sqrt{\lambda_{\min}(\ma^{\top}\ma)}}\sum_{i\in[n]}\norm{a_{i}}^{2}_2 \leq\sqrt{\kappa(\ma^{\top}\ma)}\sum_{i\in[n]}\sqrt{\sigma_{i}(\ma)}\cdot\norm{a_{i}}_{2}.
\end{equation*}
Since $\sqrt{a+b}\leq\sqrt{a}+\sqrt{b}$ this yields 
\begin{align*}
\sum_{i\in[n]} \sqrt{u_{i}} \cdot \norm{a_{i}}_2 & \leq2\sum_{i\in[n]}\sqrt{\sigma_{i}(\ma)}\cdot\norm{a_{i}}_2+\frac{1}{\sqrt{n\cdot\kappa(\ma^{\top}\ma)}}\sum_{i\in[n]}\norm{a_{i}}_{2} \leq3\sum_{i\in[n]}\sqrt{\sigma_{i}(\ma)}\cdot\norm{a_{i}}_{2}
\end{align*}
	which in turn yields the result as $\otilde$ hides factors poly-logarithmic
	in $n$ and $d$. 
\end{proof}
\section{Regression Algorithm Without Leverage Score Estimates}
\label{sec:alg_no_ls}

In the previous section we showed that we can solve regression in our desired running
time provided we have constant factor approximation
to leverage scores. Here we show how to apply this procedure repeatedly to estimate leverage scores as well. We do this by first adding a large multiple
of the identity to our matrix and then gradually decreasing this
multiple while maintaining estimates for leverage scores along the way.
This is a technique introduced in \cite{iterativeRowSampling} and
we leverage it tailored to our setting. 

A key technical ingredient for this algorithm is the following well-known
result on the reduction from leverage score computation to regression with little
overhead. Formally, Lemma \ref{lem:computing_leverage_scores} states that you can compute
constant multiplicative approximations to all leverage scores of a
matrix in nearly linear time plus the time needed to solve $\otilde(1)$
regression problems. Algorithm \ref{alg:computeLS} details the procedure for computing leverage scores. 

\begin{algorithm2e}[h]
	
	\caption{$\mathtt{ComputeLS}(\ma,\delta,\text{\ensuremath{\mathcal{A}}})$}
	
	\label{alg:computeLS}
	
	\SetAlgoLined
	
	Let $k=c\log(n)$ and $\epsilon=\frac{\delta^{2}}{(18nd\log n\cdot\kappa(\ma^{\top}\ma))^{2}}$
	where $c$ is some large enough constant.
	
	\For{$j=1,\cdots,k$}{
		
		Let $v_{j}\in\R^{n}$ be a random Gaussian vector, i.e. each entry
		follows $N(0,I)$.
		
		Use algorithm $\mathcal{A}$ to find a vector $y_{j}$ such that 
		\[
		f_{\ma,v_{j}}(y_j)-\min_{x}f_{\ma,v_{j}}(x)\leq\epsilon(f_{\ma,v_{j}}(0)-\min_{x}f_{\ma,v_{j}}(x)) ~.
		\]
	}
	
	Let $\tau_{i}=\frac{1}{k}\sum_{j=1}^{k}(e_{i}^{\top}\ma^{\top}y_{j})^{2}$
	for all $i=1,\cdots,n$.
	
	\textbf{Output:} $\frac{\tau}{1-\delta/3}+\frac{\delta}{2n\cdot\kappa(\ma^{\top}\ma)}$.
	
\end{algorithm2e}

\begin{lem}[Computing Leverage Scores]
	\label{lem:computing_leverage_scores} For $\ma\in\R^{n\times d}$
	let $\alg$ be a $\runtime(\ma)$-time algorithm for regression
	on $\ma$.
	For $\delta\in(\frac{1}{n},\frac{1}{2})$, in time
$
	O((\nnz(\ma)+\mathcal{T}(\ma) \log \epsilon^{-1} )\delta^{-2}\log n)
$
where we set
$
\epsilon=\delta^{2}(18n\cdot d\cdot\log n\cdot\kappa(\ma^{\top}\ma))^{-2},
$
	with high probability in $n$, the algorithm $\mathtt{ComputeLS}(\ma,\delta,\text{\ensuremath{\mathcal{A}}})$
	outputs $\tau\in\R^{n}$ such that 
	$
	\sigma_{i}(\ma)\leq\tau_{i}\leq(1+\delta)\sigma_{i}(\ma)+
	\delta \cdot [n\cdot\kappa(\ma^{\top}\ma)]^{-1}\,.
	$ for all $i \in [n]$.
\end{lem}

We defer the proof of Lemma \ref{lem:computing_leverage_scores} to the Appendix (Section \ref{sec:computeleveragescores}). Combining the algorithm for estimating leverage scores $\mathtt{ComputeLS}$
(Algorithm \ref{alg:computeLS}) with our regression
algorithm given leverage scores $\mathtt{SolveUsingLS}$ (Theorem~\ref{thm:solver_using_leverage_scores})
yields our solver (Algorithm~\ref{alg:mainalgo}). We first provide
a technical lemma regarding invariants maintained by the algorithm,
Lemma~\ref{lem:invariant}, and then we prove that this algorithm
has the desired properties to prove Theorem~\ref{thm:our_solvers_fine}.

\begin{algorithm2e}[h]
	
	\caption{$\mathtt{Solve}_{\ma}(x_{0},\rhs,\epsilon)$}
	
	\label{alg:mainalgo}
	
	\SetAlgoLined
	
	Let $\ma_{\eta}=\left(\begin{array}{c}
	\ma\\
	\sqrt{\eta}\mI
	\end{array}\right)$, $\eta=\lambda_{\max}(\ma^{\top}\ma)$ and $u_{i}=\begin{cases}
	\frac{1}{\eta}\norm{a_{i}}_2^{2} & \text{if }1\leq i\leq n\\
	1 & \text{if }n+1\leq i\leq n+d
	\end{cases}.$\footnote{Note that $u_{i}\in\R^{n+d}$ because $\ma_{\eta}$ has $n+d$ rows
		instead of $n$ rows.}
	
	\Repeat{$\eta>\frac{1}{10}\lambda_{\min}(\ma^{\top}\ma)$}{
		
		$u\leftarrow2\cdot\mathtt{ComputeLS}(\ma_{\eta},\frac{1}{4},\text{\ensuremath{\mathcal{A}}})$
		for algorithm $\mathcal{A}$ given by $\mathtt{SolveUsingLS}_{\ma_{\eta},u}$.
		
		$\eta\leftarrow\frac{3}{4}\cdot\eta$.
		
	}
	
	Set $\eta\leftarrow0$. Let $\overline{\rhs}=\left(\begin{array}{c}
	\rhs\\
	\vec{0}
	\end{array}\right)\in\R^{n+d}$.
	
	Apply algorithm $\mathtt{SolveUsingLS}_{\ma_{0},u}$ to find 
	$y$ such that \[
	f_{\ma_{0},\overline{\rhs}}(y)-\min_{x}f_{\ma_{0},\overline{\rhs}}(x)\leq\epsilon(f_{\ma_{0},\overline{\rhs}}(x_{0})-\min_{x}f_{\ma_{0},\overline{\rhs}}(x)) ~.
	\]
	
	\textbf{Output:} $y$
	
\end{algorithm2e}
\begin{lem}
	\label{lem:invariant} In the algorithm $\mathtt{Solve}_{\ma,\epsilon}$
	(See Algorithm~\ref{alg:mainalgo}) the following invariant is satisfied
	\begin{equation}
	\sigma_{i}(\ma_{\eta})\leq u_{i}\leq4\cdot\sigma_{i}(\ma_{\eta})+ [n\cdot\kappa(\ma_{\eta}^{\top}\ma_{\eta})]^{-1}.\label{eq:u_invariant}
	\end{equation}
\end{lem}
\begin{proof}
	Note that $\ma_{\eta}^{\top}\ma_{\eta}=\ma^{\top}\ma+\eta\mI$. Consequently,
	since initially $\eta=\lambda_{\max}(\ma^{\top}\ma)$ we have that
	initially 
	$
	\eta\mI\preceq\ma_{\eta}^{\top}\ma_{\eta}\preceq2\eta\mI$.
	Consequently, we have that initially
	$
	\sigma_{i}(\ma_{\eta})\leq u_{i}\leq2\sigma_{i}(\ma_{\eta})$
	and therefore satisfies the invariant (\ref{eq:u_invariant}).
	
	Now, suppose at the start of the repeat loop, $u$ satisfies the invariant
	(\ref{eq:u_invariant}). In this case the the assumptions needed to
	invoke $\mathtt{SolveUsingLS}$ by Theorem~\ref{thm:solver_using_leverage_scores}
	are satisfied. Hence, after the line $u\leftarrow2\cdot\mathtt{ComputeLS}(\ma_{\eta},\frac{1}{4},\text{\ensuremath{\mathcal{A}}})$,
	by Lemma~\ref{lem:computing_leverage_scores} we have that for all
	$i\in[n]$
	\[
	2\sigma_{i}(\ma_{\eta})\leq u_{i}\leq2\left(1+\frac{1}{4}\right)\sigma_{i}(\ma_{\eta})+\frac{2}{4n\cdot\kappa(\ma_{\eta}^{\top}\ma_{\eta})}\,.
	\]
	Now, letting $\eta'=\frac{3}{4}\eta$ we see that
$
	(3 /4) \sigma_{i}(\ma_{\eta})\leq\sigma_{i}(\ma_{\eta'})\leq (4 / 3) \sigma_{i}(\ma_{\eta})
$
	and direct calculation shows that invariant (\ref{eq:u_invariant})
	is still satisfied after changing $\eta$ to $\eta'$.
	
	All the remains is to consider the last step when we set $\eta=0$.
	When this happens $\eta<\frac{1}{10}\lambda_{\min}(\ma^{\top}\ma)$.
	and therefore $\sigma_{i}(\ma_{\eta})$ is
	close enough to $\sigma_{i}(\ma)$ and the invariant (\ref{eq:u_invariant})
	is satisfied.
\end{proof}
Using this we prove that our main algorithm works as desired.

\begin{proof}[Proof of Theorem~\ref{thm:our_solvers_fine}]
	Lemma~\ref{lem:invariant} shows that $u$ is always a good estimate
	enough of $\sigma_{i}(\ma_{\eta})$ throughout the algorithm to invoke
	$\mathtt{SolveUsingLS}$ with Theorem~\ref{thm:solver_using_leverage_scores}.
	In particular, this holds at the last step when $\eta$ is set to
	$0$ and thus the output of the algorithm is as desired by Theorem~\ref{thm:solver_using_leverage_scores}. 
	
	During the whole algorithm, $\mathtt{ComputeLS}(\ma_{\eta},\frac{1}{4},\text{\ensuremath{\mathcal{A}}})$
	is called $\Theta(\log(\kappa(\ma^{\top}\ma)))$ times. Each time $\mathtt{ComputeLS}$
	is called, $\mathtt{SolveUsingLS}$ is called $\Theta(\log(n))$ many
	times. All that remains is to bound the running time of $\mathtt{SolveUsingLS}$.
	However, for $\lambda\geq0$ and $i\in[n]$ we have $\sigma_{i}(\ma_{\lambda})\leq\sigma_{i}(\ma_{0})$
	and since $\ma_{\lambda}^{\top}\ma_{\lambda}\geq\lambda\mI$ we $\lambda\leq\lambda_{\min}(\ma_{\lambda}^{\top}\ma_{\lambda})$.
	Furthermore, since $\lambda_{\min}(\ma_{\lambda}^{\top}\ma_{\lambda})\geq\lambda_{\min}(\ma^{\top}\ma)$
	we have that the running time follows from the following:
\begin{equation*}
	\frac{\sum_{i\in[n+d]}\sqrt{\sigma_{i}(\ma_{\lambda})}\cdot\norm{a_{i}}_{2}}{\sqrt{\lambda_{\min}(\ma_{\lambda}^{\top}\ma_{\lambda})}} \leq\sum_{i\in[n]}\frac{\sqrt{\sigma_{i}(\ma)}\cdot\norm{a_{i}}_{2}}{\sqrt{\lambda_{\min}(\ma^{\top}\ma)}}+\sum_{i\in[d]}\frac{\sqrt{\lambda}}{\sqrt{\lambda}} ~.
\end{equation*}
\end{proof}


\section{Previous Best Running Time for Regression}
\label{sec:Deferred}

Here we state Theorem~\ref{thm:standard_solvers2}, a generalization of Theorem~\ref{thm:standard_solvers} that is useful for our analysis. Theorem~\ref{thm:standard_solvers2} follows by applying recent
results on accelerated coordinate descent \cite{accelerationNonUniform,Ns16} to the dual of regression
through recent results on approximate proximal point / Catalyst \cite{proximalPoint,catalyst}.

\begin{thm}[Previous Best Regression Running Time]
	\label{thm:standard_solvers2} Let $\ma$ and $\precond$ be matrices with the same number of columns. Suppose that $\precond$ has $n$ rows and $\left(\frac{5}{6}\right) \precond^{\top}\precond\preceq\ma^{\top}\ma\preceq
	\left(\frac{6}{5}\right) \precond^{\top}\precond$, then there is a $\mathcal{T}(\ma)$-time regression
	algorithm with
	\[
	\runtime(\ma) = 
	\otilde\left(\nnz(\ma)+\left(n + \frac{\sum_{i\in[n]}\norm{\precondrow_{i}}_{2}}{\sqrt{\lambda_{\min} (\precond^\top \precond )}}\right)\cdot s(\precond)\right).
	\]
\end{thm}

\subsection{Proof of Theorem \ref{thm:standard_solvers2}}

First we give the theorems encapsulating the results we use and then use them to prove Theorem~\ref{thm:standard_solvers}
in the case when $\ma=\precond$. We then prove the case when $\ma\neq\precond$. Theorem \ref{thm:corr_desc} describes the fastest coordinate descent algorithm known by \cite{accelerationNonUniform}. Theorem \ref{thm:proxpoint} describes the reduction \cite{proximalPoint} to from regression to coordinate decent via proximal point. 
	
\begin{thm}[{Corollary of Thm 5.1 of \cite{accelerationNonUniform}}]
	\label{thm:corr_desc} Let $f : \R^n \rightarrow \R$ be a twice differentiable
	$\sigma$-strongly convex function for $\mu > 0$. Further suppose that for all $x \in \R^{n}$ and $i\in[n]$ it is the case that $\frac{\partial^{2}}{\partial x_{i}^{2}}f(x)\leq L_{i}$
	for $i\in[n]$ and the partial derivative
	$\frac{\partial}{\partial x_{i}}f(x)$ can be computed in $O(s)$ time.
	Then there exists an algorithm which given any $\epsilon > 0$ finds a $y \in \R^n$ such that 
\[
	f(y)-\min_{x}f(x) \leq \epsilon \left(f(x_{0})-\min_{x}f(x)\right).
\]
	in \textit{expected} running time $O(s \sum_i \sqrt{L_i / \mu})$. 

\end{thm}
\begin{thm}[Corollary of Thm 4.3 of \cite{proximalPoint}]
\label{thm:proxpoint}
	Given $\ma\in\R^{n\times d}$ with rows $a_{1},...,a_{n}$ and $c\in\R^{n}$.
	Consider the function $p(x)=\sum_{i=1}^{n}\phi_{i}(a_{i}^{\top}x)$
	where $\phi_{i}$ are convex functions. Suppose that $\lambda\mI\preceq\nabla^{2}p(x)\preceq L\mI$
	for all $x\in\R^{d}$. Let $\kappa=L/\lambda$. Let dual problem $g_{s}(y)=\sum_{i=1}^{n}\phi_{i}^{*}(y_{i})+\frac{1}{2\lambda}\norm{\ma^{\top}y}_{2}^{2}-s^{\top}\ma^{\top}y$. 
	
	Suppose that for any $s\in\R^{d}$, any $y_{0}\in\R^{n}$ and any
	$0\leq\epsilon\leq\frac{1}{2}$, we can compute $y$ in expected running time $\mathcal{T}_{\epsilon}$
	such that
	\begin{equation}
	g_{s}(y)-\min_{y}g_{s}(y)\leq\epsilon(g_{s}(y_{0})-\min_{y}g_{s}(y)).\label{eq:dual_req}
	\end{equation}
	Then, for any $x_{0}$ and any $\epsilon \in (0,\frac{1}{2})$ we can find $x$ such that
	\[
	p(x)-\min_{x}p(x)\leq\epsilon\left(p(x_{0})-\min_{x}p(x)\right)
	\]
	in time $\tilde{O}(\mathcal{T}_{\delta}\log(1/\epsilon))$ w.h.p. in $n$ where $\delta=\Theta(n^{-2}\kappa^{-4})$ and $\tilde{O}$ includes logarithmic factors in $n,\kappa$. 
	\end{thm}

We note that although the guarantees of Thm 5.1 of \cite{accelerationNonUniform} and Thm 4.3 of \cite{proximalPoint} are not stated in the form of Theorems \ref{thm:corr_desc} and \ref{thm:proxpoint}. They can be easily converted to the form above by noticing that the expected running time of the procedure in Thm 4.3 of \cite{proximalPoint} using Theorem \ref{thm:corr_desc} is $\tilde{O}(T_{\delta}\log(1/\epsilon))$ which can then be boosted to high probability in $n$ using Lemma \ref{lemma:reduction}. We now give the proof of Theorem \ref{thm:standard_solvers2}.

\begin{proof}[Proof of Theorem~\ref{thm:standard_solvers2} when $\ma=\precond$]
	Let
\[
p(x)=\sum_{i=1}^{n}\phi_{i}(a_{i}^{\top}x)
\text{ where }\phi_{i}(x)=\frac{1}{2}(x-\rhs_{i})^{2} ~.
\]
	Then, we have that $\phi_{i}^{*}(y)=\frac{1}{2}y^{2}+\rhs_{i}y$ and
	hence
	\begin{equation*}
	g_{s}(y) =\sum_{i=1}^{n}\phi_{i}^{*}(y_{i})+\frac{1}{2\lambda}\norm{\ma^{\top}y}_{2}^{2}-s^{\top}\ma^{\top}y =\frac{1}{2}\norm y_2^{2}+\rhs^{\top}y+\frac{1}{2\lambda}\norm{\ma^{\top}y}_{2}^{2}-s^{T}\ma^{T}y.
	\end{equation*}
	Note that $g_{s}(y)$ is 1 strongly convex and 
	\[
	\frac{d^{2}}{dy_{i}^{2}}g_{s}(y)=1+\frac{1}{\lambda}\norm{a_{i}}_2^{2}\defeq L_{i} ~.
	\]
	Hence, Theorem~\ref{thm:corr_desc} finds $y$ satisfying \eqref{eq:dual_req}	in time
\begin{equation*}
O\left(s(\ma)\cdot\sum_{i \in [n]}\sqrt{1+\frac{1}{\lambda}\norm{a_{i}}_2^{2}}\log(\epsilon^{-1})\right) = O\left(\left(n+\frac{1}{\sqrt{\lambda}}\sum_{i\in[n]}\norm{a_{i}}_{2}\right)\cdot s(\ma)\cdot\log(\epsilon^{-1})\right).
\end{equation*}
	Hence, this shows that the primal can be solved in time 
	\[
	O\left(\left(n+\frac{1}{\sqrt{\lambda}}\sum_{i\in[n]}\norm{\precondrow_{i}}_{2}\right)\cdot s\cdot\log(n\cdot\kappa)\cdot\log(\kappa\epsilon^{-1})\right)
	\]
	where we used $\ma=\precond$ at the end.
\end{proof}
\begin{proof}[Proof of Theorem~\ref{thm:standard_solvers2} for the case $\ma\neq\precond$]
	The proof involves two steps. First, we show that given any point
	$x_{0}$, we can find a new point $x$ that is closer to the minimizer.
	Then, we bound how many steps it takes. To find $x$, we consider
	the function
	\[
	f_{x_{0}}(x)=\frac{1}{2}\norm{\precond x-\precond x_{0}}_{2}^{2}+\left\langle \ma x_{0}-c,\ma x-\ma x_{0}\right\rangle .
	\]
	Let $z$ be the minimizer of $f_{x_{0}}$ and $x^{*}$ be the minimizer
	of $\frac{1}{2}\norm{\ma x-\rhs}_{2}^{2}$. Note that 
	\begin{equation*}
	z = x_{0}-(\precond^{\top}\precond)^{-1}\ma^{\top}\eta\text{ with }\eta=\ma x_{0}-\rhs,\;\text{ and }\; x^{*} = (\ma^{\top}\ma)^{-1}\ma^{\top}\rhs.
	\end{equation*}
	Hence, we have that
	\begin{align*}
	\frac{1}{2}\norm{\ma z-\ma x^{*}}^{2}_2 & =\frac{1}{2}\norm{\ma(\ma^{\top}\ma)^{-1}\ma^{\top}\eta-\ma(\precond^{\top}\precond)^{-1}\ma^{\top}\eta}^{2}_2\\
	& =\frac{1}{2}\eta\ma^{\top}(\ma^{\top}\ma)^{-1}\ma^{\top}\eta-\eta\ma^{\top}(\precond^{\top}\precond)^{-1}\ma^{\top}\eta + \frac{1}{2}\norm{\ma(\precond^{\top}\precond)^{-1}\ma^{\top}\eta}_2^{2}.
	\end{align*}
	Using that $\frac{5}{6}\precond^{\top}\precond\preceq\ma^{\top}\ma\preceq\frac{6}{5}\precond^{\top}\precond$, 
	we have 
	\begin{equation}
	\frac{1}{2}\norm{\ma z-\ma x^{*}}_2^{2} \leq\frac{4}{10}\eta\ma^{\top}(\ma^{\top}\ma)^{-1}\ma^{\top}\eta = \frac{4}{10}\norm{\ma x_{0}-\ma x^{*}}_2^{2}.\label{eq:distAx}
	\end{equation}

	However, it is difficult to reduce to the case when $\ma=\precond$ to minimize the function
	$f_{x_{0}}$ due to the extra linear term. To address this issue, we assume $\precond=[\overline{\precond};\sqrt{\frac{\lambda}{100}}\mI]$
	by appending an extra identity term. Note that this only adds a
	small matrix $\frac{\lambda}{100}\mI$ and hence we still have $\frac{5}{6}\precond^{\top}\precond\preceq\ma^{\top}\ma\preceq\frac{6}{5}\precond^{\top}\precond$
	but with a slightly different constant which will not affect the proof
	for (\ref{eq:distAx}). Due to the extra identity term, $f_{x_{0}}(x)$ reduces to an expression of the form $\frac{1}{2}\norm{\precond x-d}_2^{2}+C$ for some vector
	$d$ and constant $C$. We can now apply Theorem \ref{thm:standard_solvers2} for the case
	$\ma=\precond$ and get an $x$ such that
	\begin{equation}
	f_{x_{0}}(x)-f_{x_{0}}(z)\leq\frac{1}{200}\left(f_{x_{0}}(x_{0})-f_{x_{0}}(z)\right).\label{eq:f_grad_gurant}
	\end{equation}
	in time 
\[
O\left(\left(n+\frac{\sum_{i\in[n]}\norm{\precondrow_{i}}_{2}}{\sqrt{\lambda_{\min}(\precond^\top \precond)}}\right)\cdot s(\precond) \cdot\log(n \kappa)\cdot\log(\kappa)\right) ~.
\]
	Note that the extra terms in $\precond$ does not affect the minimum eigenvalue
	and it increases $\frac{1}{\sqrt{\lambda}}\sum_{i\in[n]}\norm{\precondrow_{i}}_{2}$
	by atmost $n$. 
	
	Now, using the formula of $z$, the guarantee (\ref{eq:f_grad_gurant})
	can be written as 
\[
\norm{\precond x-\precond z}_{2}^{2}\leq\frac{1}{200}\norm{\ma x_{0}-\ma x^{*}}_{2}^{2} ~.
\]
	Using that $\frac{5}{6}\precond^{\top}\precond\preceq\ma^{\top}\ma\preceq\frac{6}{5}\precond^{\top}\precond$ ,
we have 
\[
\norm{\ma x-\ma z}_{2}\leq\frac{1}{10}\norm{\ma x_{0}-\ma x^{*}}_{2} ~.
\]
	Combining this with (\ref{eq:distAx}), we have that
	\[
	\norm{\ma x-\ma x^{*}}_{2}\leq0.9\norm{\ma x_{0}-\ma x^{*}}_{2}.
	\]
	Hence, we get closer to $x^{*}$ by constant factor. Therefore, to
	achieve \eqref{eq:dist_guarantee}, we only need to repeat this process
	$\log(1/\epsilon)$ times. 
Hence, the total running time is 
\[
O\left(\left(n + \frac{\sum_{i\in[n]}\norm{\precondrow_{i}}_{2}}{\sqrt{\lambda_{\min}(\precond^\top \precond)}}\right)\cdot s(\precond) \log^2(n\kappa) \log(\epsilon^{-1})\right)
\]	
\end{proof}

\section{Reduction from High Probability Solvers to Expected Running Times}

In this section we provide an auxiliary lemma that reduce the problem of achieving $\epsilon$ accuracy with high probability to the problem of achieving an accuracy $c$ with probability at least $\delta$ for some constants $c,\delta$ up to logarithmic factor blowups. Note that a naive reduction suffers an additional $\log\log(1/\epsilon)$ blowup which we avoid in the following reduction. The reduction although straightforward helps us provide a concise description of the algorithm for the ERM problem in the next section. 

\begin{lem}
\label{lemma:reduction}
Consider being given a function $F: \R^d \rightarrow \R$ and define $x^* \defeq \argmin_x F(x)$. Let $\alg$ be an algorithm such that given any point $x_0$ the algorithm runs in time $\runtime$ and produces a point $x'$ such that 
\[F(x') - F(x^*) \leq c \left( F(x_0) - F(x^*) \right)\]
with probability at least $1 - \delta$ for given universal constants $c,\delta \in [0,1]$. Further there exists a procedure $\mathcal{P}$ which given a point $x$ can produce an estimate $m$ in time $\runtime'$ such that 
\[m/r \leq F(x) - F(x^*) \leq rm\]
for some given $r \geq 1$. Then there exists a procedure that given a point $x_0$ outputs a point $x'$ such that 
\[F(x') - F(x^*) \leq \epsilon \left( F(x_0) - F(x^*) \right)\]
and the expected running time of the procedure is bounded by
\[O\left((\runtime + \runtime')\log(r)\log(\epsilon^{-1}\right))\] where $O$ hides constant factors in $c,\delta$. Moreover for any $\gamma$ we have a procedure that produces a point $x'$ such that 
\[F(x') - F(x^*) \leq \epsilon \left( F(x_0) - F(x^*) \right)\]
with probability at least $1 - \gamma$ with a total running time of 
\[O\left((\runtime + \runtime')\log(r)\log(\epsilon^{-1})\log(\gamma^{-1})\right)\]
\end{lem}
The proof of the lemma is straightforward and we defer it to the Appendix (Section \ref{sec:proofreductionlemma}).

\section{Extension for \ermname Problems}
\label{sec:erm}
In this section we consider the ERM problem defined in Definition \ref{defn:ERMdef} and prove our main result (Theorem \ref{thm:mainermtheorem}) for such problems. We propose Algorithm \ref{alg:mainalgoERM} coupled with Lemma \ref{lemma:reduction} to solve the ERM case. The algorithm takes as input estimates of leverage scores of the matrix $\ma^{\top}\ma$. The algorithm then creates an estimator of the true function by sampling component functions according to the probability distribution given by the leverage scores and appropriate re-scaling to ensure unbiasedness. Further it reformulates the estimator as a sum of variance reduced components akin to \cite{johnson2013accelerating}. The algorithm then approximately minimizes the estimator using an off-the-shelf ERM minimizer $\mathcal{A}$(in particular we use accelerated coordinate descent scheme of \cite{accelerationNonUniform}). This step can be seen as analogous to the preconditioned iteration in the case of linear regression.
 
\begin{algorithm2e}[h]
	
	\caption{$\mathtt{ERMSolve}(x_{0},\{\tau_i\}_{i = 1}^{n}, F(x) = \sum_{i = 1}^n f_i(x)), m)$}
	
	\label{alg:mainalgoERM}
	
	\SetAlgoLined
	
	Define for $k = 1 \rightarrow n$, $p_k \defeq \frac{\tau_k}{\sum_j \tau_j}$.

	Let $\mathcal{D}(j)$ be the distribution over $[1, \ldots n]$ such that $\forall\;k\;\;\Pr_{j \sim \mathcal{D}}(j = k) = p_k$

	Define for $k = 1 \rightarrow n$, $\tilde{f}_{k}(x) \defeq \frac{1}{p_{k}}\left[f_{k}(x) - \nabla f_{k}(x_0)^\top x \right] + \nabla F(x_0)^\top x $

		Sample $m$ integers $i_1 \ldots i_m \in [n]$ independently from $\mathcal{D}$.

		\If{$\sum_{t = 1}^{m} \frac{\|a_{i_t}\|_2}{\sqrt{p_{i_t}}} \leq 10 m \sum_{k = 1}^{n} \|a_{k}\|_2\sqrt{p_k}$}{Set $F_m(x) = \frac{1}{m}\sum_{t = 1}^{m} \tilde{f}_{i_t}(x)$.

		Use Theorem \ref{thm:standard_solvers_gen_ERM} to find $x'$ such that
		\[F_m(x') - \min F_m(x) \leq \frac{1}{512M^4} \left(F_m(x_0) - \min F_m(x)\right)\]
		}
	\textbf{Output:} $x'$
	
\end{algorithm2e}

Theorem \ref{thm:mainthmERM} (given below) provides the decrease guarantee and bounds the running time of Algorithm \ref{alg:mainalgoERM}. 

\begin{thm}
	\label{thm:mainthmERM}
	Given an ERM problem (Definition \ref{defn:ERMdef}) and numbers $u_i$ which are over estimates of leverage scores i.e. $u_i \geq \sigma_i$, set parameters such that
	$\tau_i = \min\{1, 20 u_i \log(d)\}$, $m = 80\left( (\sum_j \tau_j) \cdot M^4  \right)\}$ then we have that Algorithm~\ref{alg:mainalgoERM} produces a point $x'$ such that 
	\[F(x') - \min F(x) \leq \frac{1}{2}\left(F(x_0) - \min F(x) \right)\]
	with probability at least $1/2$. Further Algorithm ~\ref{alg:mainalgoERM} can be implemented in total time 
	\[\tilde{O} \left( mM + \sum_{i=1}^{n} \frac{\|a_{i}\|_2 \sqrt{\tau_i}M^3}{ \sqrt{ \lambda_{\min}(\ma^\top \ma)}} \right) ~.
	\]
\end{thm}

We now provide a quick proof sketch of Theorem \ref{thm:mainermtheorem} using Theorem \ref{thm:mainthmERM}. 

\begin{proof}[Proof of Theorem \ref{thm:mainermtheorem}]

We make use of Lemma \ref{lemma:reduction} plugging in Algorithm \ref{alg:mainalgoERM} as the procedure $\alg$. Note that $c,\delta$ are both $1/2$ as guaranteed by Theorem \ref{thm:mainthmERM}. Moreover since $F(x)$ is such that that $\forall\;\;x\;\; M\lambda_{min}(\ma^{\top}\ma) \preceq \nabla^2 F(x) \preceq M\lambda_{max}(\ma^{\top}\ma)$, we can use $\|\nabla F(x)\|_2^2$ as an estimator for $F(x) - F(x^*)$. The corresponding $r$ for it is bounded by $M^2 \kappa(\ma^{\top}\ma)$. 

Note that the running time guaranteed in Theorem \ref{thm:mainthmERM} depends on the quality of the estimates of the leverage scores input to it. We invoke Lemma \ref{lem:computing_leverage_scores} for computing accurate estimates of leverage scores. Putting together the above arguments finishes the proof for Theorem \ref{thm:mainermtheorem}.
\end{proof}
The rest of the section is devoted to proving Theorem \ref{thm:mainthmERM}. We first provide a generalization of Theorem \ref{thm:standard_solvers}.
\begin{thm}[Acc. Coordinate Descent for general ERM]
	\label{thm:standard_solvers_gen_ERM} 

	Consider the ERM problem as defined in Definition \ref{defn:ERMdef} with $\phi_i$ such that $\forall\;x\;\;\phi_i''(x) \in [\mu_i, L_i]$ and $\lambda$ such that $\forall\;x\;\nabla^2 F(x) \succeq \lambda I$. Given a point $x_0$ there exists an algorithm $\mathcal{A}$ which produces a point $x'$ w.h.p in $n$ such that
	\[
	F(x') - \min F(x^*) \leq \epsilon(F(x_0) - \min F(x^*))
	\]
	in total time proportional to 
	\[ \tilde{O} \left(\left(\sum_{i=1}^{n} \sqrt{\frac{L_i}{\mu_i}} + \sum_{i=1}^{n}\|a_i\|_2 \sqrt{\frac{L_i}{\lambda}}\right)s(\ma) \log(\epsilon^{-1}) \right)\]
\end{thm}




The proof of Theorem \ref{thm:standard_solvers_gen_ERM} is a direct consequence of \cite{accelerationNonUniform} and is deferred to the Appendix (Section \ref{sec:accproof}). We will use Algorithm $\mathcal{A}$ as a subroutine in the Algorithm \ref{alg:mainalgoERM}. 

\subsection{Proof of Theorem \ref{thm:mainthmERM}}

\begin{proof}[Proof of Theorem \ref{thm:mainthmERM}]
	For convenience we restate the definitions provided in Algorithm \ref{alg:mainalgoERM}. Given parameters $\{\tau_1 \ldots \tau_n\}$ we define a probability distribution $\mathcal{D}$ over $\{1, \ldots n\}$ such that 
	\begin{equation}
		\label{eqn:intermediatedist}
		\forall \;k \in [n]\;\;p_k \defeq Pr_{j \sim \mathcal{D}}(j = k) \defeq \frac{\tau_k}{\sum \tau_k} \quad.
	\end{equation}
	We define approximations to $f_k$ for $k \in [n]$ as 
	\begin{equation}
	\label{eqn:varredfdef}
		\tilde{f}_{k}(x) \defeq \frac{1}{p_{k}}\left[f_{k}(x) - \nabla f_{k}(x_i)^Tx \right] + \nabla F(x_i)^Tx \quad.
	\end{equation} 
	As described in the algorithm we sample $m$ integers $\{i_1, \ldots i_m\}$ independently from $\mathcal{D}$ and we define an approximation to $F$
	\begin{equation}
	\label{eqn:intermedfunc}
		F_{m}(x) \defeq \frac{1}{m}\sum_{t = 1}^{m} \tilde{f}_{i_t}(x) \quad.
	\end{equation}
	Further define 
	\[x_* = \argmin_x F(x)\]

	In order to prove the theorem we will prove two key properties. Firstly the choice of the sample size $m = \Omega(\sum_{k=1}^n \tau_k M^4)$ is sufficient to ensure that approximately minimizing $F_{m}(x)$ is enough to make constant multiplicative factor progress. Further we will bound the running time of the inner coordinate descent procedure. 

	Consider the random matrix 
	\begin{equation}
	\label{eqn:tildeaijdefn}
		\tilde{\ma}^\top\tilde{\ma} \defeq \frac{1}{m} \sum_{t=1}^{m} \frac{a_{i_t} a_{i_t}^\top}{p_{i_t}} \quad.
	\end{equation}
	Define the event $\mathcal{E}_1$ to be the following event.
	\begin{equation}
	\label{eqn:matrixconc} 
	 \mathcal{E}_1 \defeq \{ 0.5\ma^\top \ma \preceq \tilde{\ma}^\top \tilde{\ma} \preceq 2 \ma^\top \ma \}\quad.
	\end{equation}
	Via a direct application of the concentration inequality Lemma \ref{lemma:concwithreplacement} (proved in the appendix Section \ref{sec:concentrationinequality}) we have that $Pr(\mathcal{E}_1) \geq 1 - 1/d$. The following lemma shows that under the above event we get a constant factor decrease in the error if we minimized $F_m$ exactly.
	
	\begin{lem}
		\label{claim:decreaseclaim}
		Consider an ERM problem $F(x) = \sum f_i(x)$ as defined in Definition \ref{defn:ERMdef}. Let $F_{m}$ be a sample of the ERM as defined in \eqref{eqn:intermedfunc}. Let $\tilde{\ma}$ be as defined in \eqref{eqn:tildeaijdefn}. Let 
		\[ x_m \defeq \argmin_{x \in \R^d} F_m(x).\]
		Let $\mathcal{E}_1 \defeq \{ 0.5\ma^\top \ma \preceq \tilde{\ma}^\top \tilde{\ma} \preceq 2 \ma^\top \ma \}$ and let $Pr(\mathcal{E}_1) \geq p$. Then if we set $m \geq 80(\sum_j \tau_j) \cdot M^4 $, we have that 
	\begin{equation}\label{eqn:decreaseguarantee}
	\Pr \left(F(x_{m})- F(x_{*}) \leq O\left( \frac{1}{4} \left(F(x_0) - F(x_{*})\right) \right)\right) \geq p - \frac{1}{10}
	 \end{equation} 
	\end{lem}

	The above lemma bounds the number of samples required for sufficient decrease per outer iteration. For the rest of the proof we will assume that the event $\mathcal{E}_{1}$ and the property \eqref{eqn:decreaseguarantee} holds. Note by Lemma \ref{claim:decreaseclaim} that this happens with probability at least $7/10$. Further note that the probability that the condition in the if loop, i.e.
	\begin{equation}
	\label{eqn:ifloopcond}
		\sum_{t = 1}^{m} \frac{\|a_{i_t}\|_2}{\sqrt{p_{i_t}}} \leq 10 m \sum_{k = 1}^{n} \|a_{k}\|_2\sqrt{p_k}
	\end{equation}
	happens is at least $9/10$. This is a direct implication of Markov's inequality and the fact that 
	\[\E\left[\sum_{t = 1}^{m} \frac{\|a_{i_t}\|_2}{\sqrt{p_{i_t}}}\right] = m\sum_{k = 1}^{n} \|a_{k}\|_2\sqrt{p_k}\]
	Putting the above together via a simple union bound gives us that with probability at least $6/10$ all three of the following happen $\mathcal{E}_1$, Condition \eqref{eqn:decreaseguarantee} and the execution of the if loop (i.e. Condition \ref{eqn:ifloopcond} is met).

	We now show that under the above conditions we get sufficient decrease in $x'$. Firstly note that by definition we have that 
	\begin{equation}
	\label{eqn:innerguarantee}
		F_m(x') - F_m(x_m) \leq \frac{1}{512M^4}  \left( F_m(x_0) - F_m(x_m) \right) \quad.
	\end{equation}
	Note that if event $\mathcal{E}_1$ happens then
	\begin{equation}
	\label{eqn:randintermed}
	 	\forall \;x \;\;\;\frac{1}{2M}\ma^{\top}\ma \leq \hess F_{m}(x) \leq 2M\ma^{\top}\ma \quad.
	 \end{equation} 
	Now consider the RHS of \eqref{eqn:innerguarantee}
	\begin{align}
	\label{eqn:accuracycalc1}
		F_m(x_0) - F_m(x_m) &\leq M\|x_0 - x_m\|_{\ma^{\top}\ma}^2 \nonumber \\
		&\leq 2M(\|x_0 - x_*\|_{\ma^{\top}\ma}^2 + \|x_m - x_*\|_{\ma^{\top}\ma}^2) \nonumber \\
		&\leq 4M^2(F(x_0) - F(x_*) + F(x_m) - F(x_*)) \nonumber\\
		&\leq 5M^2(F(x_0) - F(x_*)) 		
	\end{align}
	The first inequality follows from Equation \ref{eqn:randintermed}, second from triangle inequality, third by noting that F is $\frac{1}{M}$ strongly convex in the norm given by $\ma^{\top}\ma$ (Assumption \eqref{eqn:generalERMEquation}) and the fourth from Lemma~\ref{claim:decreaseclaim}.
	Further note that 
	\begin{align}
	\label{eqn:accuracycalc2}
		F(x') - F(x_m) &\leq \nabla F(x_m)^{\top} (x' - x_m) + \frac{M}{2}\|x' - x_m\|^2_{\ma^{\top}\ma} \nonumber\\
		&\leq \|\nabla F(x_m)^{\top}\|_{[\ma^{\top}\ma]^{-1}}\|(x' - x_m)\|_{\ma^{\top}\ma} + \frac{M}{2}\|x' - x_m\|^2_{\ma^{\top}\ma} \nonumber\\
		&\leq \sqrt{2M(F(x_{m}) - F(x_*))}\|(x' - x_m)\|_{\ma^{\top}\ma} + \frac{M}{2}\|x' - x_m\|^2_{\ma^{\top}\ma} \nonumber\\
		& \leq \sqrt{2M(F(x_{m}) - F(x_*))}\sqrt{4M (F_{m}(x') - F_{m}(x_m))} + 2M^2(F_{m}(x') - F_{m}(x_m)) \nonumber\\ 
		&\leq \frac{1}{8M}\sqrt{(F(x_{m}) - F(x_*))}\sqrt{(F_{m}(x_0) - F_{m}(x_m))} + \frac{1}{256M^2}(F_{m}(x_0) - F_{m}(x_m)) \nonumber\\ 
		&\leq \frac{1}{3}(F(x_0) - F(x_*))
	\end{align}
	The first and third inequality follow by noting that $F$ is $M$ smooth and $1/M$ strongly convex in $\ma^{\top} \ma$ norm. Fourth inequality follows by noting that if event $\mathcal{E}_1$ holds $F_{m}$ is $1/2M$ strongly convex in $\ma^{\top} \ma$ norm. Fifth inequality follows from \eqref{eqn:innerguarantee} and sixth inequality follows from \eqref{eqn:accuracycalc1}. 

	\eqref{eqn:accuracycalc2} together with \eqref{eqn:decreaseguarantee} implies that with probability at least $6/10$, we have that 
	\begin{align*}
		F(x') - F(x_*) \leq \frac{1}{2}(F(x_0) - F(x_*))
	\end{align*}

	We will now bound the running time of the procedure via Theorem \ref{thm:standard_solvers_gen_ERM}. Define $L_{i_t}$ and $\mu_{i_t}$ to respectively the smoothness and strong convexity parameters of the components $\frac{\tilde{f}_{i_t}}{m}$.
	Note that $L_{i_t} \leq \frac{M}{m p_{i_t}}$ and $\mu_{i_t} \geq \frac{1}{M m p_{i_t}}$. Note that event $\mathcal{E}_1$ gives us that $\forall\;x\;\nabla^2 F_m(x) \succeq \frac{1}{2M}\lambda_{min}(\ma^\top \ma)$. A direct application of Theorem \ref{thm:standard_solvers_gen_ERM} using the bounds on $L_{i_t}$ and $\mu_{i_t}$ gives us that the total running time is bounded by 
	\[\tilde{O} \left(\left(\sum_{t=1}^{m} M + \sum_{t=1}^{m}\|a_{i_t}\|_2 \sqrt{\frac{M^2}{m p_{i_t}}}\right)s(\ma) \log(\epsilon^{-1}) \right) \leq \tilde{O} \left( mM + \sum_{i=1}^{n} \frac{\|a_{i}\|_2 \sqrt{\tau_i}M^3}{ \sqrt{ \lambda_{\min}(\ma^\top \ma)}} \right)\]
	The inequality follows from Condition \eqref{eqn:ifloopcond} and the definitions of $p_k,m$.

\end{proof}
\subsection{Proof of Lemma \ref{claim:decreaseclaim}}

	\begin{proof}[Proof of Lemma \ref{claim:decreaseclaim}]
	Consider the definitions in \eqref{eqn:intermediatedist}, \eqref{eqn:varredfdef}, \eqref{eqn:intermedfunc}. Note the following easy observation.
	\[ F(x) = \E_{k \sim \mathcal{D}} \tilde{f}_{k}(x)\]
	Consider the following Lemma \ref{lemma:convexfunctionbound} which connects the optima of two convex functions $F$ and $G$. 
	\begin{lem}
\label{lemma:convexfunctionbound}
Let $F(x),G(y):\R^n \rightarrow R$ be twice differentiable and strictly convex. Define
\[ x_{*} = \argmin_{x}F(x) \text{ and } y_{*} = \argmin_{y}G(y)\]
Then we have that 
\[
F(y_*)-F(x_{*}) = \norm{\grad G(x_{*})}_{\mh_G^{-1}\mh_F\mh_G^{-1}}^{2}\,.
\]
where $\mh_F \defeq \int_{0}^{1}\nabla^2 F(t.y_* + (1-t)x_*)dt$ and $\mh_G \defeq \int_0^1 \nabla^2 G(t.y_* + (1-t)x_*)$.
\end{lem}
We wish to invoke Lemma \ref{lemma:convexfunctionbound} by setting $F=F(x),G=F_m(x)$. In this setting we have that 
\[\mh_F \defeq \int_{0}^{1}\nabla^2 F(t.x_m + (1-t)x_*)dt \text{ and }\mh_G \defeq \int_0^1 \nabla^2 F_{m}(t.x_m + (1-t)x_*)\]
Firstly note that the definition of $F$ and Assumption \eqref{eqn:generalERMEquation} gives us that
\begin{equation}
\label{eqn:hesianintermedbound}
	\mh_F \preceq M \cdot \ma^{\top}\ma
\end{equation}
Using Definition \ref{eqn:tildeaijdefn} and Assumption \eqref{eqn:generalERMEquation} gives us that
\[ \mh_G \preceq M \cdot \tilde{\ma}^{\top}\tilde{\ma}\]
Combining the above two and noting that the event $\mathcal{E}_1$ happens with probability at least $p$ we get that 
\begin{equation}
\label{eqn:localintermed1}
	\mh_G^{-1} \ma^{\top}\ma \mh_G^{-1} \preceq 2M^2 [\ma^{\top}\ma]^{-1} \text{ w.p. }p
\end{equation}
Also note that for any fixed matrix $\mr$, we have that
\[ \E[\norm{\nabla F_m(x_*)}_{\mr}^2] = \frac{\E_{k \sim \mathcal{D}}[\norm{\nabla \tilde{f_{k}}(x_*)}_{\mr}^2]}{m}\]
which implies via Markov's inequality that with probability at least $9/10$ we have that 
\begin{equation}
\label{eqn:localintermed2}
	\norm{\nabla F_m(x_*)}_{\mr}^2 \leq \frac{10\E_{k \sim \mathcal{D}}[\norm{\nabla \tilde{f_{k}}(x_*)}_{\mr}^2]}{m}
\end{equation}
Putting \eqref{eqn:localintermed1} and \eqref{eqn:localintermed2} together and using a union bound we get that 
\[ \|F_{m}(x_*)\|^2_{\mh_G^{-1}\ma^{\top}\ma \mh_G^{-1}} \leq \frac{20M^2\E_{k \sim \mathcal{D}}[\norm{\nabla \tilde{f_{k}}(x_*)}_{[\ma^{\top}\ma]^{-1}}^2]}{m} \text{ w.p. } p - 1/10\]
Using Lemma \ref{lemma:convexfunctionbound} and \eqref{eqn:hesianintermedbound} we get that with probability at least $p - 1/10$
\begin{equation}
\label{eqn:mainintermed1}
F(x_m) - F(x_*) \leq \frac{20M^3\E_{k \sim \mathcal{D}}[\norm{\nabla \tilde{f_{k}}(x_*)}_{[\ma^{\top}\ma]^{-1}}^2]}{m}	
\end{equation}
We will now connect $\E_{k \sim \mathcal{D}}[\norm{\nabla \tilde{f_{k}}(x_*)}_{[\ma^{\top}\ma]^{-1}}^2]$ with the error at $x_i$. 

	\begin{lem}
	\label{lemma:ermsigma}
	Consider an ERM function $F(x) = \sum_{i=1}^{m} f_i(x)$ where $f_i(x) = \psi_i(a_i^\top x)$ with $\psi_i'' \in [\frac{1}{M}, M]$. Define a distribution $\mathcal{D}(j)$ over $[n]$ such that $Pr(j = k) = p_k \defeq \frac{\tau_k}{\sum \tau_k}$ for numbers $\tau_k \defeq \min(1, 20u_k\log(d))$ where $u_k \geq \sigma_i(\ma)$\footnote{$\sigma_i$ are leverage scores defined in Definition \ref{defn:levscores}} are overestimates of leverage scores. Given a point $\tilde{x}$ consider the variance reduced reformulation 
	\[ F(x) = \E_{k \sim D}[\tilde{f}_k(x)] \] 
	where
	\[ \tilde{f}_{k}(x) \defeq \frac{1}{p_{k}}\left[f_{k}(x) - \nabla f_{k}(\tilde{x})^\top x \right] + \nabla F(\tilde{x})^\top x \]
	Then we have that 
	\[ \E_{k \sim D} \left[ \| \nabla \tilde{f}_k(x_*)\|^2_{[\ma^{\top}\ma]^{-1}} \right] \leq 2(\sum_j \tau_j) \cdot M \cdot (F(\tilde{x}) - F(x_*)) \]
	 \end{lem} 
	Putting together \eqref{eqn:mainintermed1} and Lemma \ref{lemma:ermsigma} we get that 
	\[F(x_{m})-F(x_{*})  \leq O\left( \frac{(\sum_j \tau_j) \cdot M^4}{m} \cdot (F(x_0)-F(x_{*})) \right) \text{ w.p } p - \frac{1}{10}\]
	Lemma \ref{claim:decreaseclaim} now follows from the choice of $m$.
	\end{proof}
	We finish this section with proofs of Lemma \ref{lemma:convexfunctionbound} and \ref{lemma:ermsigma}
	\begin{proof}[Proof of Lemma \ref{lemma:convexfunctionbound}]
For all $t \in [0, 1]$ let $z(t) \defeq t\cdot y_*+(1-t)\cdot x_{*}$ for $t\in[0,1]$ and $\mh_F \defeq \int_{0}^{1}\nabla^2 F(z(t))dt$. By Taylor series expansion we have that 
\begin{align*}
F(x_{n}) & =F(x_{*})+\grad F(x_{*})^{\top}(x_{n}-x_{*})+\int_{0}^{1}\frac{1}{2}(x_{n}-x_{*})^{\top}\hess F(z(t))(x_{n}-x_{*})dt\\
 & =F(x_{*})+\frac{1}{2}  \norm{x_{n}-x_{*}}_{\mh_F}^{2}\,.
\end{align*}
Here we used that $\grad F(x_{*})=0$ and $\hess F(z(t))\succeq\mzero$
by the convexity of $F$. We also have by definition that
\[
\grad G(y_*)=\vzero
\]
and therefore
\[
\grad G(y_*)-\grad G(x_{*})=\int_{0}^{1}\hess G(z(t))(y_{*}-x_{*})\cdot dt
\]
and
\[
(y_{*}-x_{*})=-\mh^{-1}_G\grad G(x_{*})
\]
where $\mh_G \defeq \int_0^1 \nabla^2 G(z(t))$. We now have that
\begin{equation}
\label{eqn:1}
F(y_{*})-F(x_{*}) = \frac{1}{2}\norm{y_{*}-x_{*}}_{\mh_F}^{2}  = \norm{\grad G(x_{*})}_{\mh_G^{-1}\mh_F\mh_G^{-1}}^{2} 
\end{equation} 
\end{proof}
	\begin{proof}[Proof of Lemma \ref{lemma:ermsigma}]
	For the purpose of this proof it will be convenient to perform a change of basis. Define the function 
	\[  G(x) = \E_{k \sim \mathcal{D}} g_i(x) \;\text{ where }\; g_i(x) = \frac{1}{p_i}f_k((\ma^\top \ma)^{-1/2}x)\]
	Note that $G(x) = F((\ma^{\top}\ma)^{-1/2})x)$. We will first note that 
	\[
\hess g_{i}(x)=\frac{1}{p_{i}}\cdot\left[(\ma^{\top}\ma)^{-1/2}a_{i}a_{i}^{\top}(\ma^{\top}\ma)^{-1/2}\cdot\psi''_{i}(a_{i}^{\top}(\ma^{\top}\ma)^{-1/2}x)\right]
\]
and now by the cyclic property of trace and the fact that $\psi''_{i}\leq M$
we have 
\[
\tr(\hess g_{i}(x))=\frac{(\sum_j \tau_j)}{\tau_{i}}\cdot a_{i}^{\top}(\ma^{\top}\ma)^{-1}a_{i}\cdot M
\]
Note that $a_{i}^{\top}(\ma^{\top}\ma)^{-1}a_{i} \leq 1$. Now either $\tau_i = 1$ or $\tau_i = 20 a_{i}^{\top}(\ma^{\top}\ma)^{-1}a_{i} \log(d)$. In both cases we see that RHS above $\leq 1$. Therefore we get that $g_i$ is $(\sum_j \tau_j) M$ smooth. We now have the following lemma. 
\begin{lem}
\label{lemma:SVRGLemma}
	Let $\mathcal{D}$ be any distribution over $[n]$ and define $g = \E_{i \sim \mathcal{D}}[g_i(x)]$ for component convex functions $g_i$ each of which is $L$ smooth. Let $x_* \defeq \argmin\;g(x)$. We have that
	\[ \E_{i \sim D} \|\nabla g_i(x) - \nabla g_i(x_*)\|^2_2 \leq 2L(g(x) - g(x^*)) \]
\end{lem}
The proof of the above Lemma is identical to the proof of Equation 8 in \cite{johnson2013accelerating} and is provided in the appendix (Section~\ref{app:lemma:SVRG}) for completeness. Now note that 
\begin{align*}
	2(\sum_j \tau_j)M(f(\tilde{x}) - f(x_*)) &= 2(\sum_j \tau_j)\cdot M \cdot (g((\ma^\top \ma)\tilde{x}) - g((\ma^\top \ma) x_*)) \\
	&\geq \E_{i \sim \mathcal{D}} \|\nabla g_i((\ma^\top \ma)  \tilde{x}) - \nabla g_i((\ma^\top \ma) x_*)\|_2^2 \\
	&= \E_{i \sim \mathcal{D}} \|(\ma^\top \ma)^{-1/2} \frac{1}{p_i}( \nabla f_i(\tilde{x}) - \nabla f_i(x_*))\|_2^2 \\
	&= \E_{i \sim D} \| \nabla \tilde{f_i}(x_*)\|_{(\ma^\top \ma)^{-1}}^2 -  2\E_{i \sim D} \left[\frac{1}{p_i}(\nabla f_i(x_*) - \nabla f_i(\tilde{x}))^{\top} [\ma^{\top}\ma]^{-1} \nabla F(\tilde{x})\right] \\
	& \qquad\qquad\qquad\qquad\qquad\qquad\qquad\qquad\qquad\qquad\qquad\qquad - \|\nabla F(\tilde{x})\|^2_{[\ma^{\top}\ma]^{-1}} \\
	&\geq \E_{i \sim D} \| \nabla \tilde{f_i}(x_*)\|_{(\ma^\top \ma)^{-1}}^2
\end{align*}
The first line follows by definition. The second line by \ref{lemma:SVRGLemma} and by noting that $g$ is $(\sum_j \tau_j)M$ smooth. The third and fourth line follows by definition. The fifth line follows by noting that $\nabla F(x_*) = 0$. 
	\end{proof}


\bibliographystyle{alpha}
\bibliography{our_bib}

\ENDDOC

%% file: general_notation.tex
\global\long\def\R{\mathbb{R}}

\global\long\def\boldVar#1{\mathbf{#1}}
\global\long\def\mvar#1{\boldVar{#1}}
\global\long\def\vvar#1{\vec{#1}}


\global\long\def\defeq{\stackrel{\mathrm{{\scriptscriptstyle def}}}{=}}
\global\long\def\E{\mathbb{E}}
\global\long\def\otilde{\widetilde{O}}


\global\long\def\gradient{\nabla}
 \global\long\def\grad{\gradient}
 \global\long\def\hessian{\nabla^{2}}
 \global\long\def\hess{\hessian}


\newcommand{\norm}[1]{\|#1\|}



 \global\long\def\vzero{\vvar 0}

\newcommand{\ma}{\mvar A}
\newcommand{\mb}{\mvar B}

 \global\long\def\mh{\mvar H}
\global\long\def\mI{\mvar I}

 \global\long\def\mm{\mvar M}
\global\long\def\mn{\mvar N}
 
 \global\long\def\mr{\mvar R}

 \global\long\def\mzero{\mvar 0}


\global\long\def\tr{\mathrm{tr}}

\global\long\def\runtime{\mathcal{T}}

\global\long\def\argmin{\mathrm{argmin}}

\global\long\def\nnz{\mathrm{nnz}}

\newcommand{\rhs}{b}
\newcommand{\precond}{\mb}
\newcommand{\precondrow}{b}

%% file: arxiv.bbl
\newcommand{\etalchar}[1]{$^{#1}$}
\begin{thebibliography}{FGKS15b}

\bibitem[{All}16]{katyusha}
Zeyuan {Allen Zhu}.
\newblock Katyusha: Accelerated variance reduction for faster {SGD}.
\newblock {\em CoRR}, abs/1603.05953, 2016.

\bibitem[AQRY16]{accelerationNonUniform}
Zeyuan {Allen Zhu}, Zheng Qu, Peter Richt{\'{a}}rik, and Yang Yuan.
\newblock Even faster accelerated coordinate descent using non-uniform
  sampling.
\newblock In {\em Proceedings of the 33nd International Conference on Machine
  Learning, {ICML} 2016, New York City, NY, USA, June 19-24, 2016}, pages
  1110--1119, 2016.

\bibitem[CLM{\etalchar{+}}15]{uniformSampling}
Michael~B. Cohen, Yin~Tat Lee, Cameron Musco, Christopher Musco, Richard Peng,
  and Aaron Sidford.
\newblock Uniform sampling for matrix approximation.
\newblock In {\em Proceedings of the 2015 Conference on Innovations in
  Theoretical Computer Science, {ITCS} 2015, Rehovot, Israel, January 11-13,
  2015}, pages 181--190, 2015.

\bibitem[Coh16]{cohenSubspace}
Michael~B. Cohen.
\newblock Nearly tight oblivious subspace embeddings by trace inequalities.
\newblock In {\em Proceedings of the Twenty-Seventh Annual {ACM-SIAM} Symposium
  on Discrete Algorithms, {SODA} 2016, Arlington, VA, USA, January 10-12,
  2016}, pages 278--287, 2016.

\bibitem[CW13]{ClarksonW13}
Kenneth~L. Clarkson and David~P. Woodruff.
\newblock Low rank approximation and regression in input sparsity time.
\newblock In {\em Symposium on Theory of Computing Conference, STOC'13, Palo
  Alto, CA, USA, June 1-4, 2013}, pages 81--90, 2013.

\bibitem[FGKS15a]{proximalPoint}
Roy Frostig, Rong Ge, Sham Kakade, and Aaron Sidford.
\newblock Un-regularizing: approximate proximal point and faster stochastic
  algorithms for empirical risk minimization.
\newblock In {\em Proceedings of the 32nd International Conference on Machine
  Learning, {ICML} 2015, Lille, France, 6-11 July 2015}, pages 2540--2548,
  2015.

\bibitem[FGKS15b]{frostig2015competing}
Roy Frostig, Rong Ge, Sham~M Kakade, and Aaron Sidford.
\newblock Competing with the empirical risk minimizer in a single pass.
\newblock In {\em COLT}, pages 728--763, 2015.

\bibitem[Har12]{harvey2012matrix}
Nick Harvey.
\newblock Matrix concentration, 2012.

\bibitem[JZ13a]{Johnson013}
Rie Johnson and Tong Zhang.
\newblock Accelerating stochastic gradient descent using predictive variance
  reduction.
\newblock In {\em Advances in Neural Information Processing Systems 26: 27th
  Annual Conference on Neural Information Processing Systems 2013. Proceedings
  of a meeting held December 5-8, 2013, Lake Tahoe, Nevada, United States.},
  pages 315--323, 2013.

\bibitem[JZ13b]{johnson2013accelerating}
Rie Johnson and Tong Zhang.
\newblock Accelerating stochastic gradient descent using predictive variance
  reduction.
\newblock In {\em Advances in Neural Information Processing Systems}, pages
  315--323, 2013.

\bibitem[KSST09]{kakade2009applications}
S~Kakade, Shai Shalev-Shwartz, and Ambuj Tewari.
\newblock Applications of strong convexity--strong smoothness duality to
  learning with matrices.
\newblock 2009.

\bibitem[LMH15]{catalyst}
Hongzhou Lin, Julien Mairal, and Za{\"{\i}}d Harchaoui.
\newblock A universal catalyst for first-order optimization.
\newblock In {\em Advances in Neural Information Processing Systems 28: Annual
  Conference on Neural Information Processing Systems 2015, December 7-12,
  2015, Montreal, Quebec, Canada}, pages 3384--3392, 2015.

\bibitem[LMP13]{iterativeRowSampling}
Mu~Li, Gary~L. Miller, and Richard Peng.
\newblock Iterative row sampling.
\newblock In {\em 54th Annual {IEEE} Symposium on Foundations of Computer
  Science, {FOCS} 2013, 26-29 October, 2013, Berkeley, CA, {USA}}, pages
  127--136, 2013.

\bibitem[Nes83]{Nesterov1983}
Yu. Nesterov.
\newblock {A method for solving a convex programming problem with convergence
  rate 1/k\^{}2}.
\newblock {\em Doklady AN SSSR}, 269:543--547, 1983.

\bibitem[NN13]{osnap}
Jelani Nelson and Huy~L. Nguyen.
\newblock {OSNAP:} faster numerical linear algebra algorithms via sparser
  subspace embeddings.
\newblock In {\em 54th Annual {IEEE} Symposium on Foundations of Computer
  Science, {FOCS} 2013, 26-29 October, 2013, Berkeley, CA, {USA}}, pages
  117--126, 2013.

\bibitem[NS17]{Ns16}
Yurii Nesterov and Sebastian~U. Stich.
\newblock Efficiency of the accelerated coordinate descent method on structured
  optimization problems.
\newblock {\em SIAM Journal on Optimization}, 27(1):110--123, 2017.

\bibitem[SS08]{spielmanSrivastava}
Daniel~A. Spielman and Nikhil Srivastava.
\newblock Graph sparsification by effective resistances.
\newblock In {\em Proceedings of the 40th Annual {ACM} Symposium on Theory of
  Computing, Victoria, British Columbia, Canada, May 17-20, 2008}, pages
  563--568, 2008.

\bibitem[SSZ13]{sdca}
Shai Shalev-Shwartz and Tong Zhang.
\newblock Stochastic dual coordinate ascent methods for regularized loss.
\newblock {\em J. Mach. Learn. Res.}, 14(1):567--599, February 2013.

\bibitem[SZ16]{accelSDCA}
Shai Shalev{-}Shwartz and Tong Zhang.
\newblock Accelerated proximal stochastic dual coordinate ascent for
  regularized loss minimization.
\newblock {\em Math. Program.}, 155(1-2):105--145, 2016.

\bibitem[Wil12]{Williams12}
Virginia~Vassilevska Williams.
\newblock Multiplying matrices faster than coppersmith-winograd.
\newblock In {\em Proceedings of the 44th Symposium on Theory of Computing
  Conference, {STOC} 2012, New York, NY, USA, May 19 - 22, 2012}, pages
  887--898, 2012.

\bibitem[WS16]{woodworth2016tight}
Blake~E Woodworth and Nati Srebro.
\newblock Tight complexity bounds for optimizing composite objectives.
\newblock In {\em Advances in Neural Information Processing Systems}, pages
  3639--3647, 2016.

\end{thebibliography}
